\documentclass[conference]{IEEEtran}
\IEEEoverridecommandlockouts
\usepackage{cite}
\usepackage{verbatim}
\usepackage{amsmath,amssymb,amsfonts}
\usepackage{amsthm}
\usepackage{algorithm}
\usepackage{graphicx}
\usepackage{bm}
\usepackage{textcomp}
\usepackage{xcolor}
\usepackage{subcaption}
\usepackage{pstricks}
\usepackage[colorlinks=true, allcolors=blue]{hyperref}
\usepackage[utf8]{inputenc}
\usepackage{algpseudocode}
\usepackage{soul}

\def\BibTeX{{\rm B\kern-.05em{\sc i\kern-.025em b}\kern-.08em
    T\kern-.1667em\lower.7ex\hbox{E}\kern-.125emX}}

\newcommand{{\ours}}{CA-AFL}
\newtheorem{theorem}{Theorem}
\newtheorem{proposition}[theorem]{Proposition}
\theoremstyle{remark}

\begin{document}
\title{Balancing Energy Efficiency and Distributional Robustness in Over-the-Air Federated Learning}

\author{\IEEEauthorblockN{Mohamed Badi, Chaouki Ben Issaid, Anis Elgabli$^*$ and Mehdi Bennis}
\IEEEauthorblockA{Centre of
Wireless Communications (CWC),
University of Oulu, Finland \\
$^*$ King Fahd University of Petroleum and Minerals (KFUPM), Saudi Arabia\\
Email: \{Mohamed.badi, chaouki.benissaid, mehdi.bennis\}@oulu.fi, anis.elgabli@kfupm.edu.sa}
\thanks{This work was supported by the European Commission through Grant no. 101095363 (Horizon Europe SNS JU ADROIT6G project).}
}

\maketitle

\begin{abstract}
The growing number of wireless edge devices has magnified challenges concerning energy, bandwidth, latency, and data heterogeneity. These challenges have become bottlenecks for distributed learning. To address these issues, this paper presents a novel approach that ensures energy efficiency for distributionally robust federated learning (FL) with over air computation (AirComp). In this context, to effectively balance robustness with energy efficiency, we introduce a novel client selection method that integrates two complementary insights: a deterministic one that is designed for energy efficiency, and a probabilistic one designed for distributional robustness. Simulation results underscore the efficacy of the proposed algorithm, revealing its superior performance compared to baselines from both robustness and energy efficiency perspectives, achieving more than 3-fold  energy savings compared to the considered baselines.
\end{abstract}

\section{Introduction}\label{section1}
In our rapidly digitizing world, a multitude of edge devices, from smartphones to Internet of Things (IoT) systems, are continuously generating vast amounts of private data. This deluge of raw data brings about challenges in storage, capacity, and processing. Introduced in \cite{mcmahan2017communication}, federated learning (FL) proposes a solution to these challenges by allowing devices to learn collaboratively. Instead of transferring large amounts of data to a central location, devices learn a shared global model by communicating model parameters or gradients, thereby ensuring adherence to stringent privacy constraints.

However, FL comes with its own set of challenges. With a diverse range of edge devices operating in different environments, assuming an identical data distribution is often unrealistic. This diversity can result in biases, potentially favoring certain devices over others, leading to fairness concerns. To address this, the authors in \cite{mohri2019agnostic} introduced a robust min-max formulation, ensuring that the learned model performs effectively over the worst-case combination of local empirical distributions. Furthermore, a communication-efficient iterative descent-ascent algorithm was proposed to solve this problem. This approach uniquely selects only a subset of clients in each communication round. Many recent works have built upon the formulation presented in \cite{mohri2019agnostic}, adapted for both centralized and decentralized settings \cite{deng2020distributionally,  zecchin2022communication, issaid2022dr}. On the other hand, scalability and latency are among the major challenges when integrating FL into wireless edge-centric systems \cite{hellstrom2022wireless}. AirComp, with its ability to exploit the superposition property of the wireless channel, emerges as a significant solution, especially with the exponential growth of IoT devices. Several studies such as \cite{amiri2019over, amiri2020machine, elgabli2021harnessing}, have explored the use of AirComp, highlighting its communication efficiency and advantages over traditional digital schemes. Beyond scalability and latency, energy consumption is another significant concern in wireless communication systems. Transmitting data wirelessly is energy-intensive, especially for battery-powered devices, which makes efficiency in this area essential. Dynamic client scheduling has recently been considered in a number of related works. For example, \cite{du2023gradient} proposes to integrate both channel conditions and gradient norms in client selection decisions, under the mild assumptions that the maximum gradient norm and the maximum channel condition across all clients are known in advance. Yet, this approach has drawbacks, including its reliance on several tuning parameters together with the heuristic construction of its indicator. Moreover, its adaptability with scheduled clients brings about unpredictability. For a given set of tuning parameters, the expected number of scheduled clients cannot be determined because it is intrinsically tied to the characteristics of the training problem, which cannot be known in advance.
The growing emphasis on edge learning highlights the need to maximize edge devices' performance while adapting to diverse data distributions in an energy-efficient manner. Towards this goal, this work proposes a distributed learning approach that balances between energy efficiency and distributional robustness in model training.
The main contributions can be summarized as follows:
\begin{itemize}
  \item To the best of our knowledge, this is the first work to jointly address the problem of energy efficiency, and robustness to data heterogeneity, in terms of communication costs.
 \item We theoretically show that at the extremes of our configurable energy-conservation tuning parameter, the algorithm defaults either to the conventional distributionally robust AFL algorithm, which lacks channel awareness, or into a fully energy-conservative client selection algorithm. This latter approach selects clients with the lowest energy cost without performance guarantees. For intermediate values, the proposed client selection mechanism results in a blend of the two aforementioned metrics, offering a seamless transition between established algorithms and striking a balance between energy efficiency and distributional robustness performance.
  \item Our numerical results show that our proposed algorithm, coined Channel-Aware Agnostic Federated Learning ({\ours}), can achieve significant energy savings, up to one-third of the energy consumption, compared to the conventional AFL algorithm \cite{mohri2019agnostic}, with only a negligible reduction in performance. Moreover, these results demonstrate that our proposed algorithm can surpass the performance of the energy-efficient GCA algorithm \cite{du2023gradient} in terms of both worst client accuracy and global standard deviation (STD), even when the GCA algorithm's tuning parameters are experimentally configured for optimal performance.
\end{itemize}
The rest of the paper is organized as follows. Section \ref{section2} introduces the system model and the problem formulation. Section \ref{section3} details the proposed algorithm. In Section \ref{section4}, we evaluate the average and worst test accuracy of our algorithm, {\ours}, against several baselines in terms of communication rounds and total energy. Finally, the paper concludes with final remarks in Section \ref{section5}.

\section{System Model and Problem Formulation}\label{section2}
\begin{figure}[t]
\centering
\includegraphics[width=\linewidth]{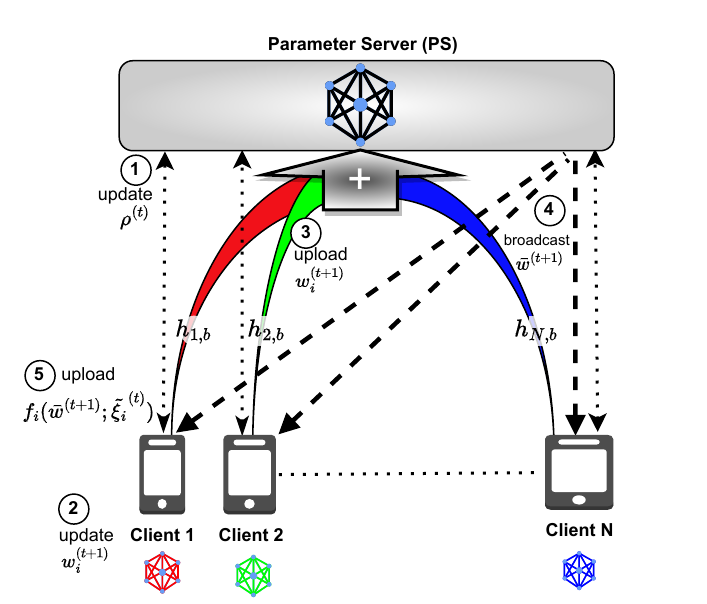}
\caption{Illustration of the AirComp System.}
\label{fig:system_model}
\end{figure}

We consider a distributed learning system as depicted in Fig.~\ref{fig:system_model}. In this system model, a parameter server (PS) functions as a central coordinator serving a total of \(N\) edge devices. Unlike digital transmission systems, where devices are required to access communication resources orthogonally to ensure successful decoding at the PS, in AirComp devices exploit analog transmission and the superposition property of the wireless channel in the uplink. This enables efficient over-the-air model aggregation. To reduce the average energy consumption, at each communication round \(t\), only \(K\) out of the \(N\) edge devices are selected to upload their models to the PS according to a probability distribution  \(\bm{\rho}^{(t)}\) defined over a combination of both energy and distributional robustness metrics. We assume that orthogonal frequency division multiplexing (OFDM) is utilized at the physical layer. Hence, leveraging AirComp the aggregated signal at the PS over the \(b\)th sub-carrier is expressed as 
\begin{align}
    y_b = \sum_{i \in D^{(t)}} h_{i,b} \times x_{i,b} + z_b,
\end{align}
where \(D^{(t)}\) is the set of sampled clients, \(x_{i,b}\) represents the data stream of the \(i\)th client on the \(b\)th sub-carrier, \(z_b\) is the additive white Gaussian noise, and \(h_{i,b}\) characterizes the channel of the \(b\)th sub-carrier between the \(i\)th client and the PS. We assume uplink-downlink channel symmetry throughout this paper.

As illustrated in Fig.~\ref{fig:system_model}, once local models are averaged at the PS, the global model is then broadcasted to all clients over a common broadcast channel. Additionally, the same figure highlights a dedicated bidirectional control channel, used to facilitate the exchange of control signals between the clients and the PS. The objective of distributionally robust optimization is to train a global model that is robust to data heterogeneity among clients. As initially discussed in \cite{mohri2019agnostic}, this optimization can be formulated as follows
\begin{align}
   \textbf{(P1)}\;\;\; \min_{\bm{w}} \max_{\bm{\lambda} \in {\Delta}^{N-1}} \left(\sum_{i=1}^{N} \lambda_{i} f_{i}(\bm{w}) \right),
    \label{min_max}
\end{align}
where \( {\Delta}^{N-1} = \{\bm{\lambda} \in \mathbb{R}^N_{+} : \sum_{i=1}^N \lambda_i = 1 \} \) is the $(N-1)$ standard simplex space. In this context, \( \bm{w} \) signifies the shared global model and \( f_i(\bm{w}) \) is the local objective function for the \(i\)th client. It is presumed that the local data for the \(i\)th client is drawn from a distribution that may vary from that of other clients. 

The energy required for the \(i\)th client, employing the channel inversion technique to upload their model to the PS is expressed as
\begin{align}
     E_i^{(t)} = P_i^{(t)} \times t_{\text{trans}},
\end{align}
where the transmission time \(t_{\text{trans}}\) is given by \(t_{\text{trans}} = (M/N_{\text{sc}})\times \tau\), with \(M\) and \(N_{\text{sc}}\) being the number of elements in the model and the number of sub-carriers, respectively, and \(\tau\) representing the symbol period. The transmission power \(P_i^{(t)}\) is defined as
\begin{align}
    P_i^{(t)} =  \sum_{b=1}^{N_{\text{sc}}} \frac{|x_{i,b}^{(t)}|^2}{|h_{i,b}^{(t)}|^2},
\end{align}
$x_{i,b}^{(t)}=\sqrt{\psi} \times s_{i,b}^{(t)}$, where $\psi$ is a scaling factor that is assumed to be constant across carriers and clients and $s_{i,b}^{(t)}$ is the actual analog value that is transmitted at $t$.  It is noteworthy that the transmission power is influenced by both the symbol power and the power necessary for channel inversion. Given that the symbol power directly reflects the learning procedure and should therefore not be a factor in the scheduling decision, our focus in this work is primarily on minimizing the energy consumption attributed to channel inversion. To that end, we define
\begin{align}
    \tilde{P}_i^{(t)} =  \psi \sum_{b=1}^{N_{\text{sc}}} \frac{1}{|h_{i,b}^{(t)}|^2}  = {\psi}\frac{N_{\text{sc}}}{|h_{i}^{(t)}|^2},
\end{align}
where the effective channel, \(h_{i}^{(t)}\),  is defined as
\begin{align}
\frac{1}{|h_{i}^{(t)}|^2}=\frac{1}{N_{\text{sc}}}\sum_{b=1}^{N_{\text{sc}}} \frac{1}{|h_{i,b}^{(t)}|^2}.    
\end{align}
Therefore, the energy attributed to the scaling and inversion operations at each transmission can be re-expressed as
\(
    \tilde{E}_i^{(t)}=\frac{\psi \times M \times \tau}{|h_{i}^{(t)}|^2}.
\) Furthermore, in this work, we denote \(E^{(t)}\) as the cumulative energy consumed by the \(K\) clients selected in the \(t\)th round for model transmission to the PS. This energy is  given by
\(E^{(t)}= \sum_{i\in D^{(t)}}{\tilde{E}_i^{(t)}}\).
\section{Proposed Algorithm}\label{section3}
The challenge of incorporating energy efficiency with distributional robustness into the client selection process arises from its inherent probabilistic nature, as proposed in the communication-efficient descent-ascent algorithms used to solve \textbf{(P1)} \cite{mohri2019agnostic, deng2020distributionally, zecchin2022communication}. In the descent step, \(K\) clients are sampled according to the probabilities defined by \(\bm{\lambda}^{(t)}\) to upload their models to the PS. However, this raises a potential issue: the selected clients might not have optimal communication channels, leading to intensive energy consumption. Hence, a refined client selection design that considers energy is crucial. With two metrics, namely the energy \(E^{(t)}\) and the probability distribution \(\bm{\lambda}^{(t)}\), our scheme transforms the deterministic energy-based client selection metric into a bias-configurable probability mass function (PMF). Governed by a tuning parameter, this allows for a smooth bias transition, which can either assign equal probabilities to all clients or give strong preference to those with the highest effective communication channels, who in turn require the least energy for model upload. 
In the following proposition, we provide a formal representation of this energy-conservative PMF.
\begin{proposition}\label{proposition1}
The entries of the bias configurable PMF can be expressed as 
\begin{align}\label{yi}
    y_i^{(t)}=\frac{\left|h_i^{(t)}\right|^C}{\sum_{j=1}^{N}{\left|h_j^{(t)}\right|^C}},
\end{align}
where \( y_i^{(t)} \) is the \(i\)-th client's sampling probability at round \( t \) and \( C \) is denoted as the energy-conservation tuning factor. 
\end{proposition}
\begin{proof}
    The proof can be found in Appendix \ref{appendix1}.
\end{proof}
 \noindent For \( C=0 \), the PMF \eqref{yi} is unbiased; as \( C \) increases, it increasingly favors clients with better channels, becoming fully biased as \( C \rightarrow \infty \). 
 
 Next, we introduce the probability distribution \(\bm{\rho}^{(t)}\), derived from the product of two distinct PMFs. At round \(t\), \(K\) clients are sampled according to
\begin{align}
    \rho_i^{(t)} = \frac{\lambda_i^{(t)}y_i^{(t)}}
    {\sum_{j=1}^{N} \lambda_j^{(t)}y_j^{(t)}}.
    \label{rho_i}
\end{align}
With this probability sampling, we aim to achieve consensus between two potentially conflicting insights. By multiplying the associated probabilities of these PMFs and subsequently normalizing the result by their summation, the weight assigned to clients evolves accordingly. Specifically, clients with elevated probabilities in both PMFs are prioritized, while clients dominant in only one PMF are given lower probabilities. Those who rank low in both PMFs are the least favored. This idea of multiplying different probability distributions and normalizing them is akin to the idea of the product of experts (PoE) in ML \cite{hinton2002training}, in which multiple probabilistic models are combined to capture complex distributions by leveraging the strengths of each individual model. In our work, we regard each PMF as an ``expert". One expert captures insights based on distributional robustness, whereas the other is geared towards energy conservation. The tuning parameter \(C\) serves a pivotal role in our algorithm, functioning as a gating mechanism. When the energy-conservative expert is unbiased, i.e., \(C \rightarrow 0\), the robustness expert takes precedence, and our algorithm defaults to the non-channel-aware AFL \cite{mohri2019agnostic}. On the other hand, when the energy efficiency expert is fully biased, i.e., \(C \rightarrow \infty\), it completely overrides the distributional robustness expert. 
For values between the two extremes, the combined PMF will represent a blend of the two insights.
Additionally, \eqref{rho_i} can be simplified using \eqref{yi} as
\begin{align}
  \rho_i^{(t)} = \frac{\lambda_i^{(t)}\left|h_i^{(t)}\right|^{C}}{\sum_{j=1}^{N} \lambda_j^{(t)}\left|h_j^{(t)}\right|^{C}}.
    \label{pit}
\end{align}
In the following proposition, we state the limiting behaviour of our algorithm as $C \rightarrow \infty$.

\begin{proposition}\label{proposition2}
    Our proposed algorithm, {\ours}, reverts to the top-$K$ greedy energy-efficient client selection as $C \rightarrow \infty$.
\end{proposition}

\begin{proof}
    The proof is deferred to Appendix \ref{appendix2}.
\end{proof}
With the definition of the client selection probability given in \eqref{pit}, our proposed algorithm operates as follows: In the descent step, \(K\) clients are sampled according to the probabilities $\bm{\rho}^{(t)}$. These clients subsequently update and upload their models using a batch of size \(|{\xi_i}^{(t)}|\). Following this, the PS broadcasts the aggregated model to all clients. On the other hand, during the ascent step, \(K\) clients are uniformly sampled to update the values of the probability simplex vector \(\bm{\lambda}^{(t)}\), employing a batch of size \(|\tilde{\xi_i}^{(t)}|\). Notably, because it is a scalar, this update does not demand a high communication cost and can be conveniently exchanged with the PS over a control channel as shown in Fig.~\ref{fig:system_model}. The detailed steps are summarized in Algorithm \ref{algo}. 

\begin{algorithm}[t]
\caption{Channel-Aware Agnostic Federated Learning (CA-AFL)}
\label{algo}
\begin{algorithmic}[1]

\Require \(N\) Clients, \(K\) sampled clients, \(T\) total number of iterations, energy-conservation factor \(C\) , learning rates (\(\eta, \gamma\)), initial model \(\bar{\bm{w}}^{(0)}\), and initial \({\bm{\lambda}}^{(0)}\).
\Ensure \(\bar{\bm{w}}^{(T)}\), \({\bm{\lambda}}^{(T)}\)

\For{\(t = 0\) to \(T-1\)}
    \State PS updates \(\bm{\rho}^{(t)}\) according to \eqref{pit}
    \State PS samples \(K\) clients \(D^{(t)} \) according to \(\bm{\rho}^{(t)}\)
    \For{clients \(i \in D^{(t)}\)} in parallel
        \State \({\bm{w_i}}^{(t+1)} = {\bm{w_i}}^{(t)} - \eta \nabla f_i({\bm{w_i}}^{(t)}; \xi_i^{(t)})\)
    \State client uploads \({\bm{w_i}}^{(t+1)}\) to the PS
    \EndFor
    \State PS averages the AirComp aggregated models
    
    \begin{align}
        \bar{\bm{w}}^{(t+1)}=\frac{\sum_{i \in D^{(t)}} {\bm{w_i}}^{(t+1)}+z^{(t)}}{K}
        \label{Fedavg}
    \end{align}
    
    \State PS broadcasts \(\bar{\bm{w}}^{(t+1)}\) to all clients
    \State PS samples \(K\) clients uniformly \(U^{(t)}\)
    \For{clients \(i \in U^{(t)}\)} in parallel
        \State compute and upload \(f_i(\bar{\bm{w}}^{(t+1)}; \tilde{\xi_i}^{(t)})\) to the PS
        \State PS updates \(\tilde{\bm{\lambda}}^{(t+1)}\) according to 
            \[
                \tilde{\lambda_i}^{(t+1)}=\lambda_i^{(t)}+\gamma f_i(\bar{\bm{w}}^{(t+1)}; \tilde{\xi_i}^{(t)})
                \label{ascent_step}
            \]
    \EndFor
    \State PS computes \(\bm{\lambda}^{(t+1)} =\Pi_{{\Delta}} (\tilde{\bm{\lambda}}^{(t+1)})\)
\EndFor
\end{algorithmic}
\end{algorithm}
\section{Numerical Results}\label{section4}
\subsection{Simulation Setup}
In this section, we detail the performance of our proposed algorithm, \ours, and evaluate its effectiveness compared to other existing algorithms. In our experiments, we utilized the Fashion MNIST dataset and deployed a logistic regression model of size \(M=7850\). Our setup includes a total of \(N=100\) devices, out of which \(K=40\) actively participate in each communication round. We sorted the 60000 training data samples by label, then divided them into 100 equal-sized shards. This approach ensures that each client receives one shard, promoting a high degree of data heterogeneity among clients. 
The training process spanned \(T=500\) rounds with a batch size of 50. For the descent step, we use an exponentially decaying learning rate with initial value \(\eta^{(0)}=0.1\) and a decaying rate of 0.998 per iteration. The ascent step size was set to \(\gamma=8 \times 10^{-3}\).
Regarding communication, for a fair comparison against the non-channel-aware AFL algorithm in terms of energy consumption, we did not impose any power control mechanism. Instead, we adopted an independent and identically distributed (i.i.d.) truncated flat-fading Rayleigh distributed channel block \(\mathcal{CN}(0,1)\), where \(h \ge 0.05\). This setup favored the non-channel-aware baseline AFL. In the most challenging scenario, we assumed the channel remained coherent for only one communication round. The scaling factor was set at \(\psi=0.5\) mW, and the symbol period, \(\tau\), aligned with LTE standards at 1 ms.
We averaged the results over five simulation runs. For the purpose of comparison, we consider the following baselines
\begin{itemize}
    \item \textbf{FedAvg} \cite{mcmahan2017communication}: The original algorithm that is neither distributionally robust nor channel-aware. In this approach, \(K\) clients are randomly sampled to upload their models to the parameter server for averaging.
    \item \textbf{AFL} \cite{mohri2019agnostic}: A distributionally robust but non-channel-aware algorithm designed for solving the min-max formulation. In each round, \(K\) clients are sampled based on \(\lambda^{(t)}\) for the descent steps, and a separate set of \(K\) clients is sampled uniformly for the ascent step update.
    \item \textbf{GCA} \cite{du2023gradient}: A non-robust, gradient and channel-aware algorithm that dynamically schedules clients based on a composite indicator consisting of energy, channel condition, and gradient norm. Its tuning parameters are \(\lambda_E=0.5\), \(\lambda_V=0.5\), \(\rho_1=0.5\), \(\rho_2=0.5\), following \cite{du2023gradient}, while \(\sigma_t=1\) and \(\alpha=1500\) are experimentally optimized, resulting in an average of \(42\) scheduled clients.
\end{itemize}
\subsection{Results and Discussion}
\begin{figure*}[t]
\centering
\begin{subfigure}{.3\textwidth}
  \centering
  \includegraphics[scale= 0.7]{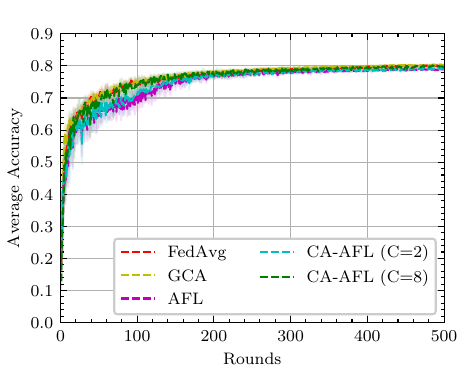}  
  \caption{Average test accuracy}
  \label{fig:avgvsrounds}
\end{subfigure}
\begin{subfigure}{.3\textwidth}
  \centering
  \includegraphics[scale=0.7]{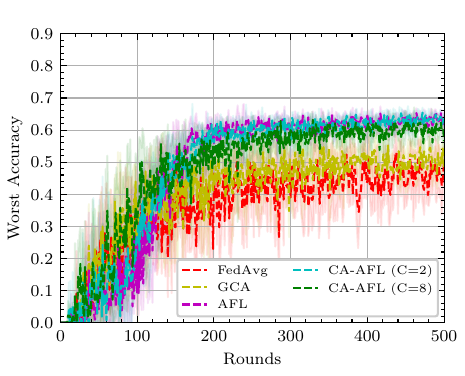} 
  \caption{Worst test accuracy}
  \label{fig:worstvsrounds}
\end{subfigure}
\begin{subfigure}{.3\textwidth}
  \centering
  \includegraphics[scale=0.7]{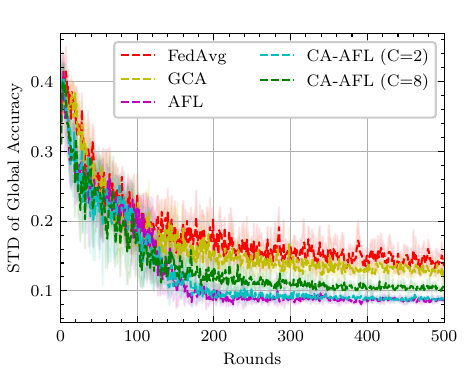} 
  \caption{STD}
  \label{fig:stdvsrounds}
\end{subfigure}
\caption{Performance of {\ours} compared to baselines in terms of the number of communication rounds.} 
\label{fig:vsrounds}
\end{figure*}
\begin{figure*}[t]
\centering
\begin{subfigure}{.3\textwidth}
  \centering
  \includegraphics[scale=0.7]{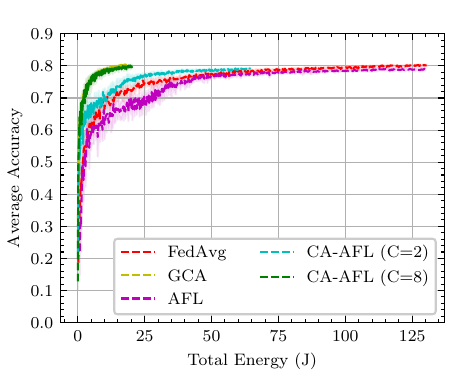}  
  \caption{Average test accuracy}
  \label{fig:avgvsenergy}
\end{subfigure}
\begin{subfigure}{.3\textwidth}
  \centering
  \includegraphics[scale=0.7]{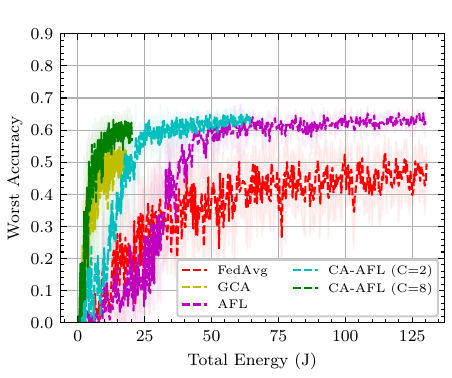} 
  \caption{Worst test accuracy}
  \label{fig:worstvsenergy}
\end{subfigure}
\begin{subfigure}{.3\textwidth}
  \centering
  \includegraphics[scale=0.7]{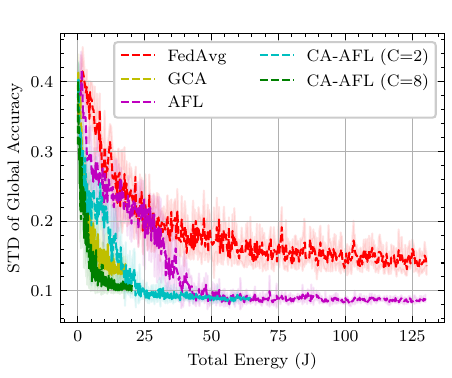} 
  \caption{STD}
  \label{fig:stdvsenergy}
\end{subfigure}
\caption{Performance of {\ours} compared to baselines in terms of the total energy.} 
\label{fig:vsenergy}
\end{figure*}
In Fig.~\ref{fig:vsrounds}, we depict the performance metrics in terms of average global accuracy, worst client accuracy, and the STD of the global accuracy versus the number of communication rounds. As shown in Fig.~\ref{fig:avgvsrounds}, the proposed {\ours} together with the baselines exhibit comparable average global accuracy with a convergence value of \(80\%\).
 Fig.~\ref{fig:worstvsrounds} illustrates that {\ours} closely matches the performance of the non-channel-aware benchmark AFL algorithm in terms of worst client accuracy, with only a negligible degradation for \(C=8\). It delivers approximately \(10\%\) higher worst accuracy than both FedAvg and the GCA algorithm. Notably, {\ours} attains the \(50\%\) worst accuracy, which is the maximum achieved by GCA and FedAvg, in less than half of the communication rounds compared to both FedAvg and GCA. 
Fig.~\ref{fig:stdvsrounds} underscores that for varying values of the tuning parameter \(C\), {\ours} outperforms both GCA and FedAvg in terms of STD, which directly translates into more reliability in terms of fairness. {\ours} aligns closely with AFL's STD for \(C=2\) and remains competitive for \(C=8\). Similar to the worst accuracy metric, {\ours} reaches the convergence STD values of FedAvg and GCA baselines with significantly fewer communication rounds. The superior performance of GCA compared to FedAvg is attributed to incorporating user gradient norms in the client selection mechanism, which accelerates convergence. However, as expected, increasing the value of \(C\) leads to a degradation in terms of both worst-client accuracy and STD compared to the robust benchmark AFL.

In Fig.~\ref{fig:vsenergy}, we plot the accuracy performance metrics, including average accuracy, worst client accuracy, and the STD of global accuracy, against the total energy expenditure for uploading. As presented in Fig.~\ref{fig:avgvsenergy}, FedAvg and AFL algorithms emerge as the most energy-intensive. This aligns with our expectations since neither of them is channel-aware. The figure further shows that increasing \(C\) for the proposed {\ours} leads to comparable average accuracy results but at considerably reduced energy costs. Notably, at \(C=8\), {\ours} attains an energy efficiency that is on par with GCA algorithm without compromising performance. Fig.~\ref{fig:worstvsenergy} underscores that {\ours} delivers performance metrics in line with the robustness benchmark but with significantly reduced energy demands. Remarkably, at \(C=8\), {\ours} not only matches the energy efficiency of GCA, it also achieves the best attainable worst client accuracy for GCA and FedAvg with substantially less energy consumption. Fig.~\ref{fig:stdvsenergy} shows that {\ours} outperforms the robustness benchmark AFL in energy efficiency, incurring only a minimal performance trade-off. At \(C=8\), {\ours} surpasses GCA in terms of STD while consuming the same amount of energy, even with GCA operating at its optimal hyperparameter configuration. On the other hand, it is obvious that {\ours} achieves the best attainable STD of GCA with much lower energy. An intriguing observation is that the proposed CA-AFL can match the performance of the benchmark AFL algorithm while consuming only one-third of the energy.

\section{Conclusion}\label{section5}
This work introduced a novel approach to incorporate energy efficiency considerations into the clients' selection process governed by the solution of the distributionally robust min-max formulation. Our proposed algorithm leverages a tuning configurable parameter, transitioning between two well-established algorithms---AFL and top-\(K\) greedy selection. Through extensive simulations, we have demonstrated that our algorithm outperforms existing baselines, in terms of both energy efficiency and distributional robustness.
\bibliographystyle{IEEEtran}
\bibliography{references} 
\section{Appendices}
\subsection{Proof of Proposition \ref{proposition1}}\label{appendix1}
We define \(G: \mathbb{R}_{+}^N \to \Delta^{N-1}\) as a transformation that projects an \(N\)-dimensional vector of non-negative real numbers onto an \((N-1)\)-dimensional probability simplex. Let \(\bm{x} = [x_{1}, x_{2}, \ldots, x_{N}]\) be a vector in \(\mathbb{R}_{+}^N\) and \(\bm{y} = [y_{1}, y_{2}, \ldots, y_{N}]\) be a vector in \(\Delta^{N-1}\). Additionally, we consider the symmetric group \(S_N\), which represents all possible permutations of \(N\) distinct elements. The transformation \(G\) adheres to the following properties:
\begin{itemize}
\item \textbf{Permutation equivariance:} The transformation \(G\) is equivariant under the action of the symmetric group \(S_N\) on \(\bm{x}\). In other words, any permutation of the inputs \(\bm{x}\) will yield an equivalent permutation of the outputs \(\bm{y}\).
    \item \textbf{Permutation invariance for non-corresponding inputs:} For any given output \(y_i\), \(G\) is invariant under the action of the permutation group \(S_{N-1}\) on the set of inputs \(\{x_j : j \in \{1, \ldots, N\}\) and \(j \neq i\}\). 
\end{itemize}
These properties allow us to express \(y_i\) as a function of \(x_i\) and a permutation-invariant function \(\zeta\) acting on the vector \(\bm{x}_{-i}\), where \(\bm{x}_{-i}\) excludes the \(i\)th element of \(\bm{x}\). The function is given by
\(
y_i = \phi(x_i, \zeta(\bm{x}_{-i}))
\). Furthermore, if \(x_i > x_j\), then \(\phi(x_i, \zeta(\bm{x}_{-i})) > \phi(x_j, \zeta(\bm{x}_{-j}))\) is also upheld, owing to the \textbf{order preservation} property.
A potential representation of the function \(\phi(x_i, \zeta(\bm{x}_{-i}))\), that satisfies the aforementioned properties, is
\begin{align}
\phi(x_i, \zeta(\bm{x}_{-i})) = \frac{g(x_i)}{\sum_{j=1}^{N} g(x_j)},
\end{align}
where \(\zeta(\bm{x}_{-i})=\sum_{j=1, j\ne i}^{N} {g({x_j})}\), and \(g\) is a monotonically increasing function, defined on the domain of non-negative real numbers and exhibits well-defined derivatives across its domain.
The function \(g\) is characterized primarily by its non-zero Taylor series coefficients
\(
g(x) = \sum_{k=1}^{M} a_k x^{C_k}
\). Given this characterization, each \(y_i\) can be formulated as
\begin{align}
y_i = \frac{\sum_{k=1}^{M} a_k x_i^{C_k}}{\sum_{j=1}^{N} \sum_{k=1}^{M} a_k x_j^{C_k}}.
\end{align}
An essential aspect of \(G\) is its reactivity to the elements in \(\bm{x}\). We term \(G\) ``unbiased`` by the values of \(\bm{x}\)" when \(y_i = y_j\), for all indices \(i\) and \(j\). This unbiased behavior emerges when all exponents \(C_k\) are zero.
In stark contrast, \(G\) is dubbed ``fully biased" if \(y_m = 1\), whenever \(x_m\) exceeds all other components of \(\bm{x}\), i.e., \(x_m > x_i\) \( \forall i \neq m\). This is satisfied when \(\max_k\{C_k\}\) tends to infinity and all other \(C_k\) tend to zero. Consequently, we represent \(g(x)\) as \(g(x) = (L(x))^C\), designating it as our bias function where \(L\) is monotonic chosen as \(L(x) = x\).
\subsection{Proof of Proposition \ref{proposition2}}\label{appendix2}
Let \( \bm{\rho}^{[k]} \) be the probability vector for the \( k \)th client's sampling procedure at round \( t \), where \( \bm{\rho}^{[k]} \in {\Delta}^{N-1} \). Without loss of generality, let \( \bm{\rho}^{[1]} = [\rho_m^{[1]}, \rho_{m-1}^{[1]}, \dots, \rho_1^{[1]}] \) be sorted in descending order with respect to the values of the effective channels, i.e., \( h_1 < \cdots < h_{m-1} < h_m \). For the particular case of \( \rho_i^{[1]} \), we have
\begin{align}
\lim_{C \rightarrow \infty} \rho_i^{[1]} = \lim_{C \rightarrow \infty} \frac{\lambda_i|h_i|^C}{\sum_{j=1}^{N} \lambda_j|h_j|^C}.
\label{limit_rhoi}
\end{align}
Furthermore, we can deduce
\begin{align}
\lim_{C \rightarrow \infty} \frac{\rho_i^{[1]}}{\rho_j^{[1]}} & =  \lim_{C \rightarrow \infty} \frac{\lambda_i}{\lambda_j}\left(\frac{|h_i|}{|h_j|}\right)^C.
\end{align}
From these results, we derive
\begin{align}\label{limratio}
\lim_{C \rightarrow \infty} \frac{\rho_i^{[1]}}{\rho_j^{[1]}} & =\left\{
    \begin{array}{ll}
           0,     & \text{ if } i < j,\\
           1,                              &\text{ if } i = j.
         \end{array}
    \right.
\end{align}
Using \eqref{limit_rhoi}, we further express
\begin{align}\label{rhoi_induction}
\lim_{C \rightarrow \infty} \rho_i^{[1]} = \lim_{C \rightarrow \infty} \frac{\rho_i^{[1]}/\rho_m^{[1]}}{\sum_{j=1}^{N} \rho_j^{[1]}/\rho_m^{[1]}} =\left\{
    \begin{array}{ll}
           0,     & \text{ if } i \neq m,\\
           1,                              &\text{ if } i = m.
         \end{array}
    \right.
\end{align}
This suggests that as \( C \) increases, our algorithm tends to select the client that requires the lowest energy for model upload. Further exploration reveals
\begin{align}
\rho_i^{[2]} & = \sum_{j=1,\; j \neq i}^{N} \frac{\rho_j^{[1]}}{1 - \rho_j^{[1]}} \rho_i^{[1]}.
\end{align}
Consequently, we have
\begin{align}
\lim_{{C \to \infty}} \rho_i^{[2]} = \left\{
    \begin{array}{ll}
          \lim\limits_{C \to \infty} \frac{\rho_m^{[1]}}{1 - \rho_m^{[1]}} \rho_i^{[1]}, & \text{if } i \neq m, \\
            0, & \text{if } i = m.
         \end{array}
    \right.
\end{align}
These insights indicate a dimensionality reduction within the simplex space. This can be articulated by the following relationships for $i \neq m$
\begin{align}
\lim_{C \rightarrow \infty} \rho_i^{[2]} & = \lim_{C \rightarrow \infty} \frac{\rho_i^{[1]}}{\rho_{m-1}^{[1]} + \dots + \rho_1^{[1]}}.
\end{align}
Furthermore, if $i \neq m$ and $j \neq m$, then we can write
\begin{align}
\lim_{C \rightarrow \infty} \rho_i^{[2]} & = \lim_{C \rightarrow \infty} \frac{\rho_i^{[1]}/\rho_{m-1}^{[1]}}{\sum_{j=1}^{N} \rho_j^{[1]}/\rho_{m-1}^{[1]}}.
 \label{pi_2}
\end{align}
Using \eqref{limratio} in \eqref{pi_2}, we deduce
\begin{align}
\lim_{C \rightarrow \infty} \frac{\rho_i^{[2]}}{\rho_j^{[2]}} =\left\{
    \begin{array}{ll}
           0,     & \text{ if } i < j, i \neq m , j \neq m,\\
           1,                              &\text{ if } i = j, i \neq m.
         \end{array}
    \right.
\end{align}
By upholding the prior conditions using mathematical induction, we establish that  as \( C \to \infty \), our algorithm simplifies to a greedy strategy, prioritizing the \( K \) clients who require the least energy to transmit their models to the PS at round \( t \).
\end{document}